\documentclass[10pt,twocolumn,letterpaper]{article}

\usepackage{cvpr}              

\usepackage{graphicx}
\usepackage{amsmath}
\usepackage{amssymb}
\usepackage{booktabs}
\usepackage{multirow}
\usepackage{bbding}
\usepackage{algorithm}
\usepackage{algorithmic}
\usepackage{bm}
\usepackage{booktabs}
\usepackage[table,xcdraw]{xcolor}
\usepackage[pagebackref,breaklinks,colorlinks]{hyperref}
\usepackage[toc,title,page]{appendix}

\usepackage[capitalize]{cleveref}
\crefname{section}{Sec.}{Secs.}
\Crefname{section}{Section}{Sections}
\Crefname{table}{Table}{Tables}
\crefname{table}{Tab.}{Tabs.}

\def\cvprPaperID{1781} 
\def\confName{CVPR}
\def\confYear{2023}

\begin{document}

\title{DUAW: Data-free Universal Adversarial Watermark against Stable Diffusion Customization}

\author{Xiaoyu Ye\footnotemark[1]\\
Peking University\\
{\tt\small yexiaoyu0711@stu.pku.edu.cn}
\and
Hao Huang\footnotemark[1]\\
Peking University\\
{\tt\small huanghao@stu.pku.edu.cn}
\and
Jiaqi An\\
Peking University\\
{\tt\small jq.an@outlook.com}
\and
Yongtao Wang\footnotemark[2]\\
Peking University\\
{\tt\small wyt@pku.edu.cn}}

\maketitle

\renewcommand{\thefootnote}{\fnsymbol{footnote}} 
\footnotetext[1]{These authors contributed equally to this work.} 
\footnotetext[2]{Corresponding author.}

\begin{abstract}
Stable Diffusion (SD) customization approaches enable users to personalize SD model outputs, greatly enhancing the flexibility and diversity of AI art. However, they also allow individuals to plagiarize specific styles or subjects from copyrighted images, which raises significant concerns about potential copyright infringement.
To address this issue, we propose an invisible data-free universal adversarial watermark (DUAW), aiming to protect a myriad of copyrighted images from different customization approaches across various versions of SD models. First, DUAW is designed to disrupt the variational autoencoder during SD customization. 
Second, DUAW operates in a data-free context, where it is trained on synthetic images produced by a Large Language Model (LLM) and a pretrained SD model. This approach circumvents the necessity of directly handling copyrighted images, thereby preserving their confidentiality. Once crafted, DUAW can be imperceptibly integrated into massive copyrighted images, serving as a protective measure by inducing significant distortions in the images generated by customized SD models. Experimental results demonstrate that DUAW can effectively distort the outputs of fine-tuned SD models, rendering them discernible to both human observers and a simple classifier.
\end{abstract}
\vspace{-5mm}
\section{Introduction}
\label{sec:introduction}
\vspace{+2mm}


\begin{figure}[t]
  \centering
  \includegraphics[width=8cm]{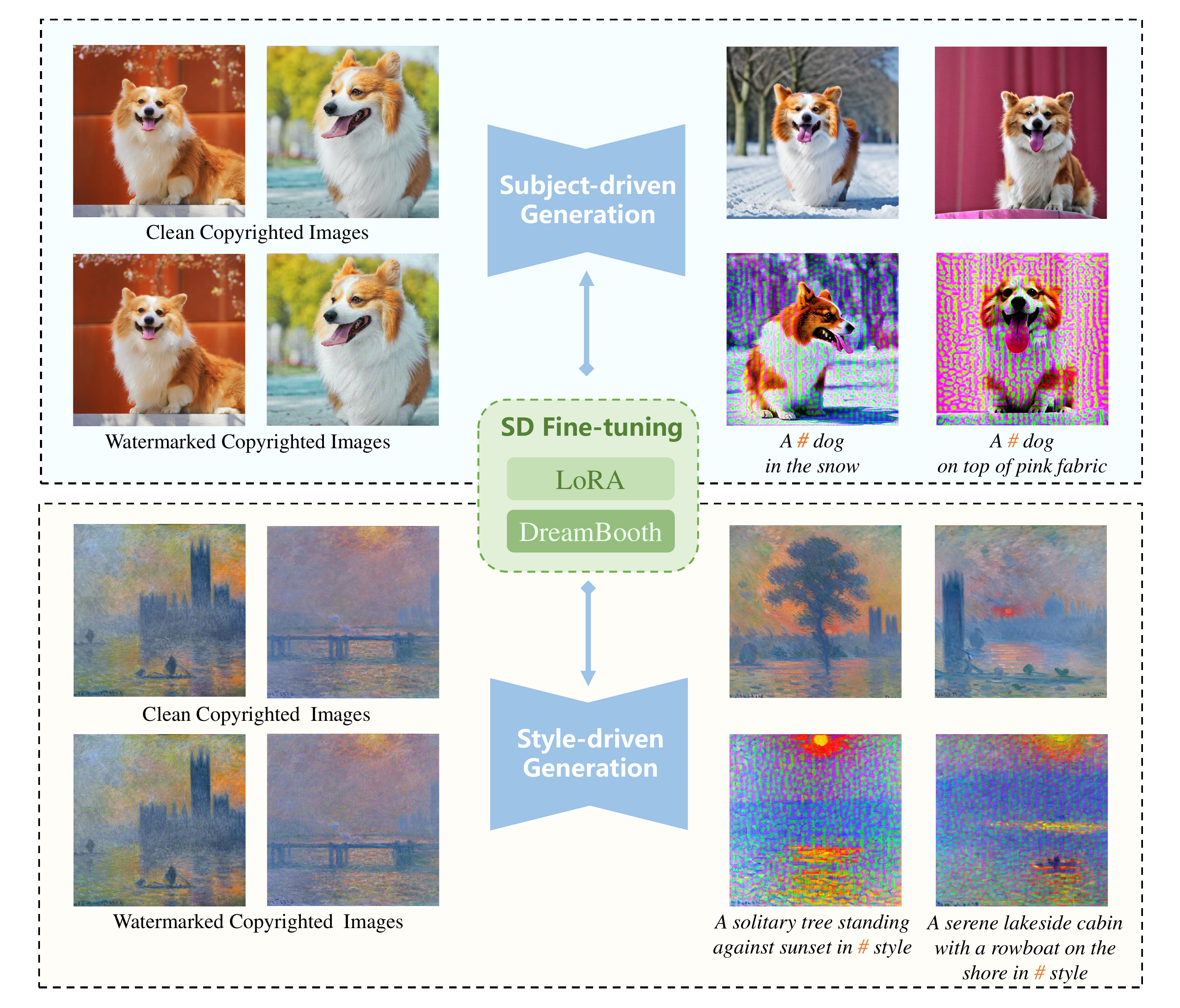}
  \vspace{-2mm}
  \caption{Illustrations of Using DUAW to Achieve Copyright Protection against SD Customization. 
  The proposed DUAW can disrupt the SD fine-tuning process, resulting in the fine-tuned model generating images with distortions.
  }
  \vspace{-4mm}
  \label{fig:demo}
\end{figure}

\begin{figure*}[t]
  \centering
  \includegraphics[width=17cm]{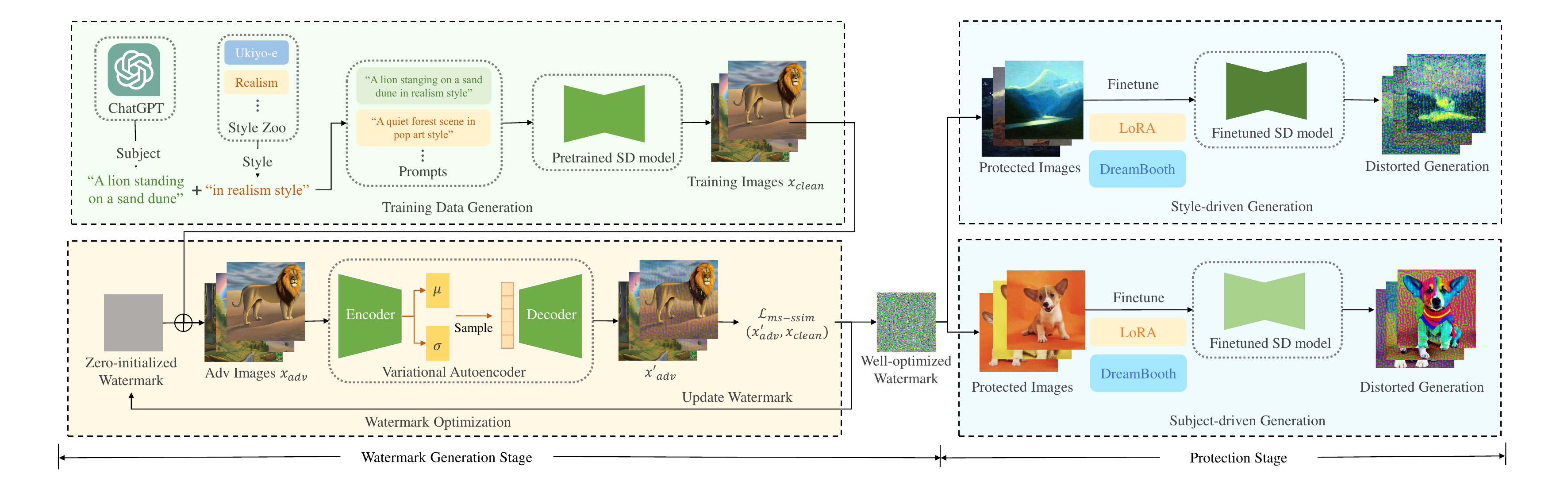}
  \vspace{-2mm}
  \caption{Overall Pipeline of DUAW. The overall process can be divided into two stages: 1) watermark generation stage and 2) protection stage. At stage 1, we first synthesize training images via LLM and pretrained SD model, and then we optimize the randomly initialized watermark by disrupting the VAE of SD models. At stage 2, the DUAW can be added to the copyrighted images, which can interfer with the SD customization process and induce obvious distortion in generated images.}
  \vspace{-2mm}
  
  \label{fig:the_main_pipeline}
\end{figure*}
\vspace{-3mm}

Recently, images generated by Stable Diffusion (SD)~\cite{rombach2022high} have exhibited an exceptional level of visual quality, exerting a profound influence on the academic and industrial communities. The open-source community has also developed several SD customization tools, \emph{e.g.}, DreamBooth~\cite{ruiz2023dreambooth} and LoRA~\cite{hu2022lora}, which enable users to personalize SD models and generate images according to their preferences. However, these tools have raised concerns about potential intellectual property (IP) infringement risks, as individuals can conveniently customize their SD models to plagiarize a specific subject or style from copyrighted images. Hence, how to protect copyrighted images from these SD customization approaches is an important issue that needs to be addressed.

In this paper, we propose to use a universal adversarial watermark to achieve copyright protection against common SD customization approaches (as shown in Fig.~\ref{fig:demo}). However, achieving this goal is non-trivial. 
Firstly, unlike conventional adversarial attacks that aim to maximize loss for a fixed model, disrupting SD customization approaches is challenging as the weight of the target model will change dynamically to minimize loss, and we have to maximize on top of the minimization process.
Secondly, Stability AI has released several SD models with different weights (such as SD v1.4, SD v1.5, and SD v2.1), highlighting the need for the proposed watermark to generalize across various SD models. Thirdly, due to the potentially high level of confidentiality associated with copyrighted images, we may not have direct access to the images that need to be protected when training the adversarial watermark.


To address the mentioned concerns, we propose a novel data-free adversarial watermark generation framework for copyright protection. This approach is based on two observations: a) Mainstream SD models, despite variations in model weights and customization methods, incorporate a variational autoencoder (VAE)~\cite{kingma2022autoencoding} for image encoding and decoding, with its parameters frozen during customization; b) Text-to-image customization techniques like DreamBooth and LoRA tend to preserve and amplify minor perturbations present in training images within generated images (Fig.~\ref{fig:keep_perturbation}). Leveraging the second observation, if we add meticulously crafted perturbations to training images, these perturbations will be maintained and magnified in generated images; Moreover, if these perturbations can disrupt the VAE, the fine-tuned SD model will produce significantly distorted images, which aligns with our objectives.

Specifically, we adopt an optimizer-based adversarial watermark training approach, with the optimization objective of minimizing similarity between a range of diverse training images $x\in \mathcal{X}$ and the images $\hat{x}$ reconstructed by the VAE with our watermark implemented, ensuring universal protection across different images. Moreover, we propose a data-free method to generate the training images based on ChatGPT and SD v2.1, which can be used to train the proposed invisible adversarial watermark without directly accessing the protected images. With just 100 synthetic images, the generated DUAW can successfully protect a wide range of images from being utilized by the SD model in subject-and-style-driven generation.

    

Our contributions can be summarized as the following:
\begin{itemize}
    \item We are the first to introduce a data-free universal adversarial watermark (DUAW) generation framework with the goal of distorting the outputs of customized SD models, effectively protecting copyrighted subjects or styles across diverse images, models, and SD customization techniques.
    
    \item We propose a simple yet effective method to craft the proposed watermark by disrupting the VAE of SD models, thus avoiding performance degradation caused by weight alterations of SD customization.
    \item We introduce a data-free solution that utilizes LLM and SD models to generate training data, eliminating the necessity of directly accessing highly confidential copyrighted images.
    \item The experimental results demonstrate that the proposed DUAW can cause obvious distortion to the outputs of fine-tuned SD models, with 96.43\% of output images successfully identified as infringing images by a simple classifier.
\end{itemize}

\section{Related Work}
\label{sec:related_work}

\subsection{Customized Text-to-Image Generation}

While diffusion-based generators~\cite{rombach2022high} have already demonstrated their unparalleled capabilities in text-to-image generation, the story of SD model customization just begins. Textual Inversion~\cite{gal2022image}, the first attempt to the best of our knowledge, learns a new subject or style in the embedding space and uses the learned new embedding for personalized generation. However, since it does not change the parameters of SD models, its potential for high-fidelity generation is constrained. DreamBooth~\cite{ruiz2023dreambooth} emerges as a more promising method, as it fine-tunes the SD model to learn an association between the subject or style and a distinctive rare text token. This association enables the generation of outputs with exceptional fidelity. LoRA~\cite{hu2022lora} is another widely-used technique, which involves adding additional rank-decomposition matrices into the attention layers of the models to customize diffusion models efficiently.


In this work, we focus on defending DreamBooth and LoRA, the two most commonly used methods by the community, with the objective of safeguarding copyrighted images from being utilized in SD customization (\emph{i.e.}, subject-driven and style-driven generation).

\subsection{Adversarial Attacks}

Adversarial attack is initially proposed in image classification~\cite{szegedy2014intriguing}, wherein imperceptible perturbations are meticulously crafted and incorporated into clean images (\emph{i.e.}, adversarial examples), effectively duping deep neural networks into producing erroneous predictions. Recently, the exploration of adversarial examples has emerged as a research hotspot, and researchers are continually exploring them in various domains, such as object detection~\cite{huang2023t, huang2021rpattack}, video recognition~\cite{jiang2023efficient} and large language models~\cite{zou2023universal}. Meanwhile, to further explore the mechanism of adversarial examples, some studies try to improve the image-level and model-level transferability of adversarial examples. UAP~\cite{moosavi2017universal} fools a recognition model
with only a single adversarial perturbation rather than crafting image-specific perturbations. CMUA-Watermark~\cite{huang2022cmua} protects massive face images from multiple deepfake models with only one adversarial watermark.

In this work, akin to prior studies, we likewise employ an adversarial watermark with high transferability across different tasks, models, and input images. However, we are unable to freeze the model weights as the classical setting of adversarial attacks,  thereby rendering it a more difficult task.








\subsection{IP Protection against SD Customization}

SD customization methods like DreamBooth and LoRA can learn a specific representation of a subject or style, posing the potential risks of copyright infringement. To mitigate this threat, existing works utilize adversarial attack to disrupt the diffusion process or encoder of target SD models and craft image-specific protection watermarks. AdvDM~\cite{liang2023adversarial}, Anti-DreamBooth~\cite{vanle2023antidreambooth}, and UDP~\cite{zhao2023unlearnable} protect copyrighted images by optimizing adversarial watermarks to disrupt the diffusion process, preventing the representation learning process. Glaze~\cite{shan2023glaze} focuses on attacking the encoder of SD models to force the encoder to produce incorrect features, thereby safeguarding artists from style imitation. Recently, ADAF~\cite{wu2023promptrobust} devises effective attack strategies targeting both the encoder and diffusion process.

Different from the previous works, the proposed DUAW can simultaneously protect massive copyrighted images from different SD models and customization approaches without direct access to these images.

\begin{figure}[t]
  \centering
  \includegraphics[width=7cm]{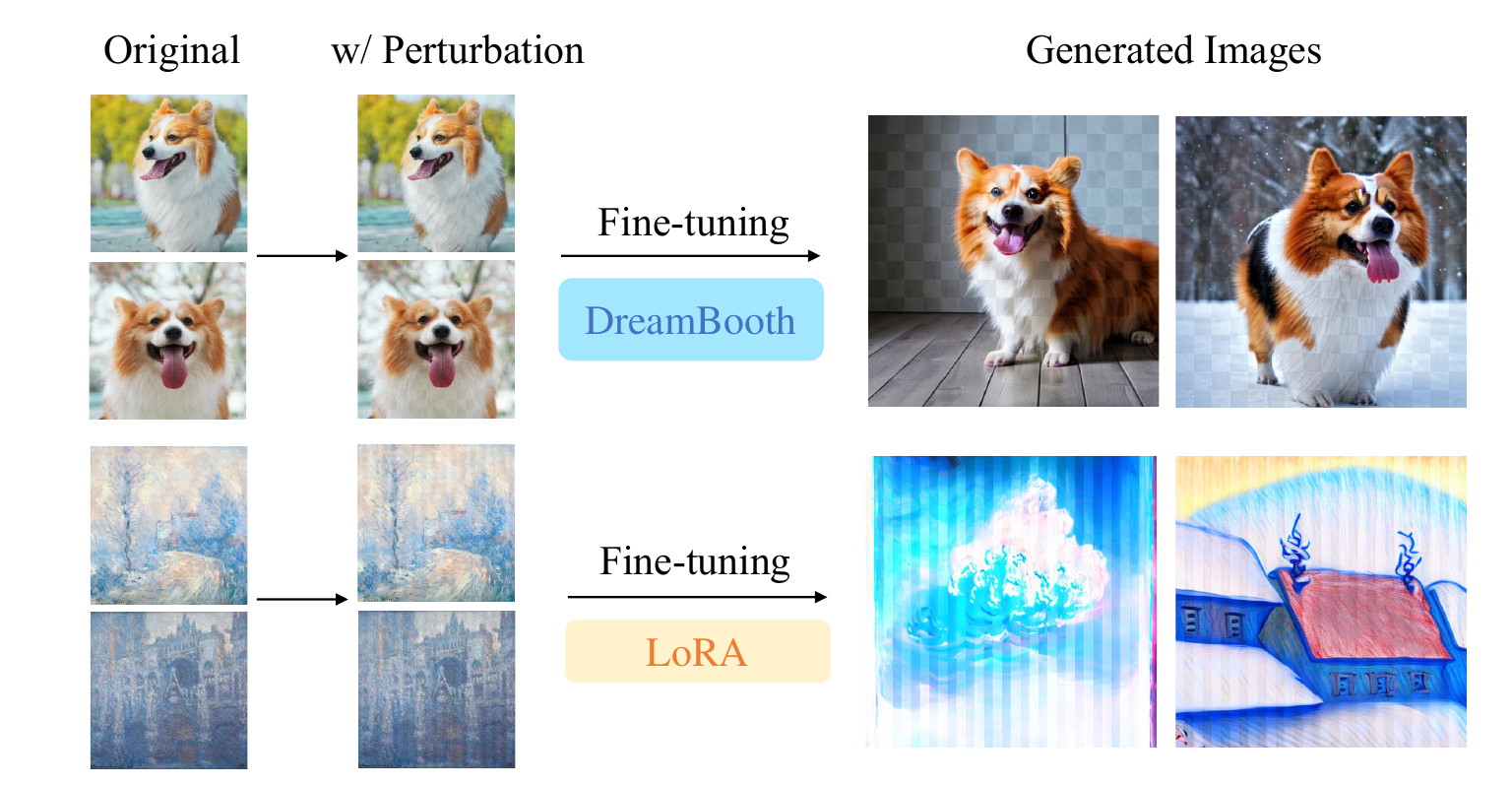}
  \vspace{-2mm}
  \caption{Examples of Fine-tuned SD Models Preserving Minor Perturbations. We apply tiny checkerboard-like and stripe perturbations with a small upper bound (0.03) to the training images. After fine-tuning, both DreamBooth and LoRA obviously preserve and accentuate the pre-added perturbations in the generated images.
  }
  \vspace{-4mm}
  \label{fig:keep_perturbation}
\end{figure}

\vspace{-3mm}

\section{Method}
\label{sec:method}



\subsection{Background}
\label{sec:Background}
\paragraph{Stable Diffusion}
Diffusion models~\cite{sohl2015deep} learn the dataset distribution $p(x)$ by reconstructing images progressively blurred with Gaussian noise $\alpha$ through denoising autoencoders $\alpha_\theta(x_t,t);t = 1\ldots T$. 
As a prompt-based variant, Stable Diffusion (SD)~\cite{rombach2022high} operates in the latent space of a pretrained VAE. Specifically, it learns the conditional distribution $p(z|y)$, where $z$ represents the latent code of $x$ obtained from the VAE encoder $\mathcal{E}$, and $y$ denotes input prompts (\emph{e.g.}, text prompts, images) which will be encoded by a domain-specific encoder $\tau_\theta$. In summary, the optimization objective of the SD model is to precisely estimate the introduced Gaussian noise:
\begin{equation}
\label{equation:SD}
    \mathcal{L}_{SD}=\mathbb{E}_{\mathcal{E}(x), y, \alpha \sim \mathcal{N}(0,1), t}\left[\left\|\alpha-\alpha_\theta\left(z_t, t, \tau_\theta(y)\right)\right\|_2^2\right],
\end{equation}
where $\tau_\theta$ and $\alpha_\theta$ of the SD model would be trained. We focus on text-to-image SD models in this work, which means $y$ would be text prompts and $\tau_\theta$ would be text encoders.

\paragraph{SD Customization}

LoRA~\cite{hu2022lora} and DreamBooth~\cite{ruiz2023dreambooth} are proposed to customize the contents in the generated
results using a small set of images with the same styles or objects. Specifically, DreamBooth links instance features in the image to a unique identifier and introduces a class-specific prior preservation loss to retain the model's prior knowledge of instance classes. The objective of DreamBooth combines both the reconstruction loss (Eq.~\ref{equation:SD}) and the prior preservation loss:
\begin{equation}
    \begin{aligned}
\mathcal{L}_{DB} = & \mathbb{E}_{\mathcal{E}(x), y, \alpha,\alpha_{pr}, t, t_{pr}}\left\|\alpha-\alpha_\theta\left(z_{t}, t, \tau_\theta(y)\right)\right\|_2^2 \\
&+\lambda\left\|\alpha_{pr}-\alpha_{\theta_{\text {ori }}}\left(z_{t_{pr}}, t_{pr}, \tau_\theta(y_{pr})\right)\right\|_2^2,
\end{aligned}
\end{equation}
where $\lambda$ is the weight for prior preservation loss and $y_{pr}$ represents the prompts for class images. Meanwhile, LoRA utilizes the standard SD model's reconstruction loss (Eq.~\ref{equation:SD}) and introduces rank-decomposition weight matrices (update matrices) to fine-tune model weights as follows:
\begin{equation}
\begin{aligned}
    W^{\prime}= & W_0+B A, \\
    & B \in \mathbb{R}^{d \times r}, A \in \mathbb{R}^{r \times k},
\end{aligned}
\end{equation}
where $W_0$ is the original weight of SD model, $W^{\prime}$ the fine-tuned, and the rank $r \ll \min(d,k)$.Since these weight matrices are lightweight and plug-and-play, LoRA significantly reduces the cost of fine-tuning SD models.

\subsection{Problem Definition}
\label{sec:problem definition}


SD customization approaches raise concerns regarding potential copyright infringement. To address this issue, we hope to propose an adversarial protection watermark to disrupt the SD fine-tuning process and induce distortion to output images generated by customized SD. Meanwhile, to mitigate the threat of different customization approaches employed by different models on various training images, the proposed watermark must exhibit generalizability to consistently protect copyrighted images under these circumstances. Hence, we formally define the following: for any set of images requiring protection, denoted as $x\in \mathcal{X}$, we apply our proposed watermark $W$ to each image, resulting in protected images $x_{adv} = \{x^{(i)} + W \}_i^n$. These watermarked images are then used as inputs to the pretrained SD model $S_\theta$ with parameters $\theta$, which undergoes fine-tuning using the fine-tuning tool $f$. 
Our objective is to induce the fine-tuned SD model to produce distorted images with any prompts $y\in \mathcal{Y}$ by minimizing a certain image quality metric $\mathcal{Q}$:

\begin{equation}
\label{equation:main}
\begin{aligned}
 &W  = \text{arg}\min_W \mathbb{E}_{x,y}(\mathcal{Q}(S_{\theta^{\prime}}(y))),\\
 &\theta^{\prime} = f(S_{\theta},x_{adv}),\\
 &s.t.\|W\|_{\infty} < \epsilon,
\end{aligned}
\end{equation}
where $\theta^{\prime}$ is the fine-tuned weight of SD model, $y$ is any prompt to condition the SD generation, and $\epsilon$ is a predefined upper bound that limits the maximum pixel alteration of the watermark $W$.

\subsection{Overview of Pipeline}
\label{sec:overview}

As depicted in Fig.~\ref{fig:the_main_pipeline}, our pipeline can be divided into watermark generation stage and protection stage. During the watermark generation stage, we leverage diverse painting styles in combination with painting content prompts provided by ChatGPT and feed the prompts to a pretrained SD model to generate training data. Subsequently, we apply our watermark to the training data and optimize it by minimizing  Multi-scale Structural Similarity (MS-SSIM)~\cite{wang2003multiscale} between outputs images and original images, introducing obvious distortions in the decoder's output. In the protection stage, we apply the well-optimized DUAW to any image requiring protection. When the SD model undergoes customization using DreamBooth or LoRA, the output of the SD model will be distorted, also leading to a decrease in image quality.

\subsection{Watermark Generation Process}
\label{sec:Watermark Generation on Encoder-decoder}


Directly optimizing the objective in Eq.~\ref{equation:SD} presents challenges as we must optimize the loss on top of a dynamic weight-changing process, which differs from the conventional adversarial attack setting where the target model weight is frozen. To tackle this issue, we set our eyes on the part of the SD model that stays unchanged during the fine-tuning process, \emph{i.e.}, VAE, and focus on disrupting it. Meanwhile, we have also observed that SD model fine-tuning methods tend to preserve minor perturbations added to training images (shown in Fig.~\ref{fig:keep_perturbation}). That is to say, if our watermark $W$ can perturb the decoding process of the VAE and result in distorted output, the fine-tuning process will also let the SD model learn such a special distribution of latent codes that can disrupt the decoder and introduce distortions into output images. 



Specifically, during the watermark training, the encoder of the VAE maps the watermarked images $x_{adv}$ to the parameters of the posterior distribution $q(z_{adv}|x_{adv})$ over the latent code $z_{adv}$. VAE approximates the $q(z_{adv}|x_{adv})$ using a Gaussian distribution, where the mean and standard deviation are given by the VAE encoder $\mathcal{E}$:
\begin{equation}
\label{equation:encoder}
\begin{aligned}
&\mu_{adv}, \sigma_{adv} = \mathcal{E}(x_{adv}), \\
& z_{adv} \sim \mathcal{N}(\mu_{adv}, \sigma_{adv}).
\end{aligned}
\end{equation}
After that, the decoder of the VAE maps $z_{adv}$ back to the data space to produce reconstructed Image $\hat{x_{adv}} = \mathcal{D}(z_{adv})$. To induce distortions and lower the image quality of decoder outputs, we minimize the MS-SSIM between $x$ and $\hat{x_{adv}}$. MS-SSIM is an image quality assessment metric that measures the structural similarity between two images at multiple scales. In contrast to the commonly employed Mean Square Error (MSE), MS-SSIM simulates various viewing angles and distances, offering a more comprehensive assessment of the generated image quality:
\begin{equation}
\label{equation:ms-ssim}
    \mathcal{MS}(x, \hat{x_{adv}}) = \prod_{i=1}^N \text{SSIM}(x^{(i)}, \hat{x_{adv}}^{(i)}))^{\alpha^{(i)}},
\end{equation}
where $x^{(i)}$ and $\hat{x_{adv}}^{(i)}$ is original and reconstructed images at $i$th scale, $N$ is the number of scales, $\text{SSIM}$ is the structural similarity index~\cite{wang2004image} at the $i$th scale, and $\alpha^{(i)}$ is the weight for each scale. Based on MS-SSIM (Eq.~\ref{equation:ms-ssim}), our optimization objective is:
\begin{equation}
\label{equation:ms-ssim_with_expectation}
\begin{aligned}
     \mathcal{L}_{ms-ssim} = \mathbb{E}_{x, z_{adv} \sim \mathcal{N}(\mu_{adv}, \sigma_{adv})} \mathcal{MS}(x, \mathcal{D}(z_{adv})).\\
\end{aligned}
\end{equation}
The watermark training process takes batches of images as input and updates the watermark across the entire dataset, ensuring that the generated watermark is capable of protecting multiple images simultaneously.

Due to the difficulty of directly optimizing the expectation over $z_{adv}$, we approximate it by sampling a single $z$ for each $x$. VAE provides a reparametrization trick to sample $z_{adv}$, which includes sampling latent variables $\zeta$ from a standard Gaussian distribution and transforming them into the desired distribution with the mean and standard deviation given by the encoder. Hence, our optimization objective can be rewritten as:
\begin{equation}
\label{equqation:reparam}
\begin{aligned}
    z_{adv} &= \mu_{adv} \cdot \zeta + \sigma_{adv}, \ \zeta \sim \mathcal{N}(0,1),\\
    \mathcal{L}_{ms-ssim} &= \mathbb{E}_x \mathcal{MS}(x, \mathcal{D}(z_{adv})).\\
\end{aligned}
\end{equation}

\begin{algorithm}[t]
\caption{DUAW Generation} 
\label{algorithm:watermark_generation}
\begin{algorithmic}[1]

\REQUIRE ~~
 $\mathcal{X}$ (training images), $\mathcal{E}$ (the VAE encoder), $\mathcal{D}$ (the VAE decoder), $e$ (training epochs)

\ENSURE ~~
DUAW watermark $W_{duaw}$

\STATE Zero Init $W_0$
\FOR{$i \in [0,e)$}
    \FOR{ $x \in \mathcal{X}$}
        \STATE $z_{adv}$ $\Leftarrow$ Sample from $\mathcal{E}(x + W_i)$
        \STATE $\mathcal{L}_{ms-ssim}$ $\Leftarrow$ $\mathcal{MS}(x, \mathcal{D}(z_{adv}))$
        \STATE $W_{i+1}$ $\Leftarrow$ Optimize with $\mathcal{L}_{ms-ssim}$
        \ENDFOR
\ENDFOR
\STATE $W_{duaw} = W_e$

\end{algorithmic}
\end{algorithm}

In conclusion, the complete process of watermark generation is outlined in Algorithm ~\ref{algorithm:watermark_generation}. 
\begin{figure*}[!ht]
  \centering
  \includegraphics[width=17cm]{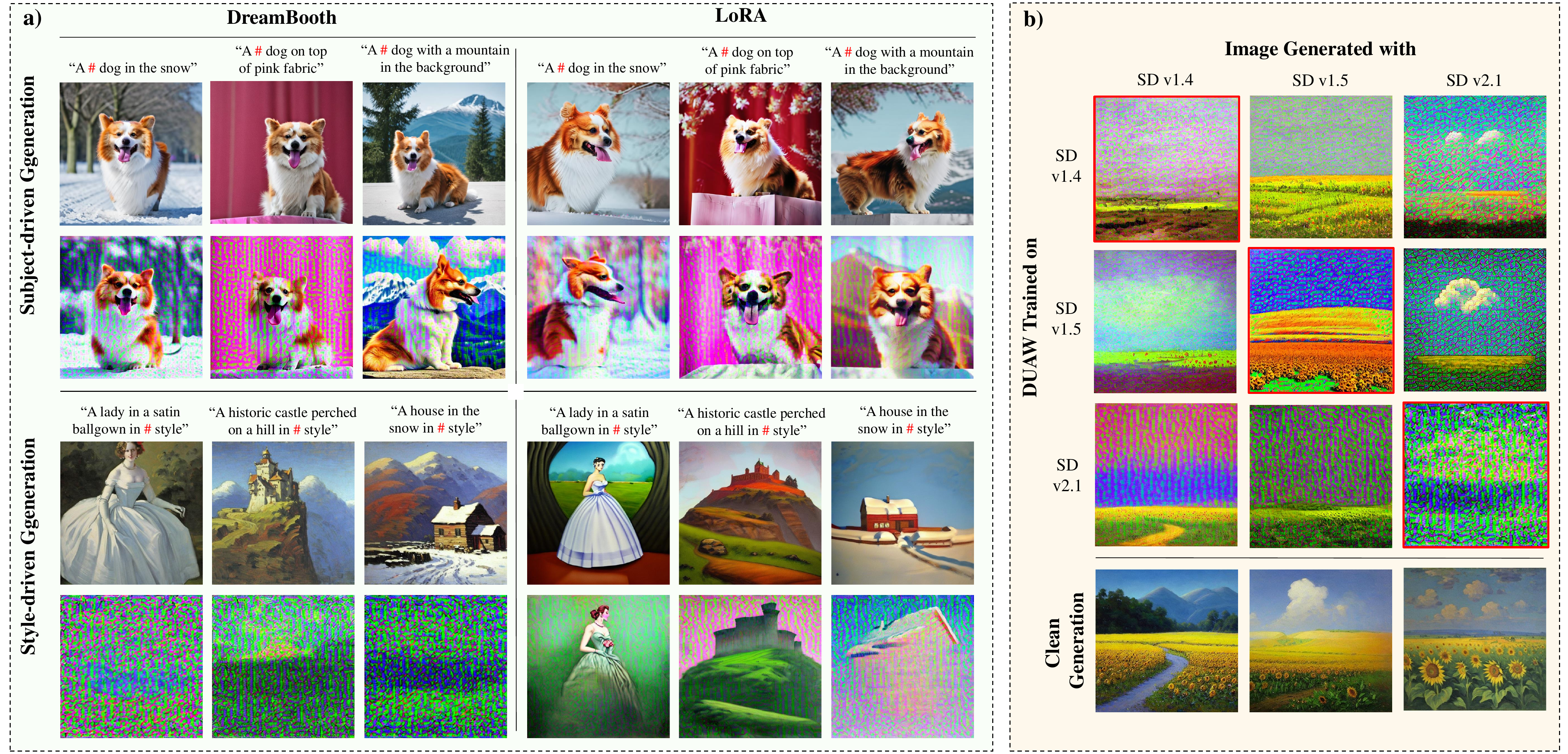}
  \vspace{-2mm}
  \caption{Qualitative Results of DUAW on DreamBooth and WikiArt Datasets.
  a) The result of the proposed DUAW on SD v2.1. The 1st and 3rd rows show images of customized SD models trained on clean images; The 2nd and 4th rows show images of customized SD models trained on watermarked images (\emph{i.e.}, clean images with DUAW). b) The transferability results on different SD versions. Each row is applied with DUAW trained on a specific version of SD, and each column is generated by a specific SD. The clean generation w/o DUAW is also provided for comparison. All images are prompted by ``A field of sunflowers in \# style''. }
  \vspace{-4mm}
  \label{fig:wikiart_db}
\end{figure*}

\subsection{Data-free Synthetic Dataset via LLM}
\label{sec:Data-free Synthetic Dataset via LLM}


In certain situations, the valuable copyrighted images needing protection might not be accessible for watermark training. For example, undisclosed characters and styles may be at risk of disclosure if used for training. Therefore, protection without accessing copyrighted images is essential.
Fortunately, we observed that the proposed watermark's protection effectiveness does not strongly depend on the training dataset, \emph{e.g.}, a watermark trained on a dataset for subject-driven generation tasks also exhibits protection capabilities on style-driven task datasets. This observation leads us to consider the use of an SD-generated dataset as a substitute for the original dataset.

We choose to utilize a pretrained SD model to create the surrogate dataset. pretrained SD model allows us to effortlessly generate diverse images, strengthening the watermark's protection against various image types, \emph{i.e.}, ensuring our watermark is universal.
We generate our surrogate dataset as follows: we prompt an LLM to give prompts about painting contents $c \in \mathcal{C}$. To improve diversity in the training data, we combine each prompt with various painting styles from the style zoo $s \in \mathcal{S}$ to form the final prompt. We then input this prompt into the SD model $S_\theta$ to obtain the training data $x$:
\begin{equation}
    x = S_\theta(c \cdot s),
\end{equation}
where $\cdot$ is the concatenation operator. Experimental results show that our watermark, trained on the surrogate dataset, delivers commendable protection performance with a small dataset size of only 100 synthetic images.


\section{Experiment}
\label{sec:experiment}

\subsection{Implementation Details} 

\paragraph{Evaluation Datasets}

For subject-driven generation, we adopt DreamBooth dataset~\cite{ruiz2023dreambooth}, which consists of 30 subject classes (21 objects and 9 live objects) with $3\sim 5$ images per class. For style-driven generation, we select 12 artists with diverse styles from the WikiArt~\footnote{https://www.wikiart.org/}, each contributing 10 artworks that are closely aligned in terms of artistic style and time period. Each image is resized and center-cropped to a resolution of $512\times 512$.

\paragraph{Training Datasets}

We first employ ChatGPT 3.5~\footnote{ChatGPT (March 23 version). https://chat.openai.com} to generate 10 random painting content and combine each content with 10 distinct painting styles to form diverse input prompts. Then, we utilize SD v2.1 to generate a 512×512 image for each prompt, resulting in 100 training images.

\paragraph{Watermark Optimization}
We set the size of our watermark to 512×512 and constrain the perturbation value within the range of $[-0.05, 0.05]$. During the training process, we set the batch size to 4, total epochs to 1000 and use Adam~\cite{kingma2017adam} as the optimizer. To regulate the learning rate, we apply the learning rate scheduler proposed by T-SEA~\cite{huang2023t}, starting with an initial learning rate of 0.01.



\begin{table*}[!ht]
\centering
\scalebox{0.9}{

\begin{tabular}{c|cccc|cccc}
\toprule
                                             & \multicolumn{4}{c|}{\textbf{WikiArt Dataset}}               & \multicolumn{4}{c}{\textbf{DreamBooth Dataset}}\\
                                             & \multicolumn{1}{c}{}                          & CLIP-Score$\downarrow$                            & IL-NIQE$\uparrow$                             & \multicolumn{1}{c|}{}                           & \multicolumn{1}{c}{}                          & CLIP-Score$\downarrow$                            & IL-NIQE$\uparrow$                             & \multicolumn{1}{c}{}                           \\
    \multirow{-3}{*}{\textbf{Method}} & \multicolumn{1}{c}{\multirow{-2}{*}{Version}} & Clean/ Adv                            & Clean/ Adv                          & \multicolumn{1}{c|}{\multirow{-2}{*}{SR (\%)$\uparrow$}} & \multicolumn{1}{c}{\multirow{-2}{*}{Version}} & Clean/ Adv                            & Clean/ Adv                          & \multicolumn{1}{c}{\multirow{-2}{*}{SR (\%)$\uparrow$}} \\
\midrule
& \cellcolor[HTML]{EFEFEF}v1.4                  & \cellcolor[HTML]{EFEFEF}0.5828/ 0.5681 & \cellcolor[HTML]{EFEFEF}35.62/ 43.82 & \cellcolor[HTML]{EFEFEF}94.34                  & \cellcolor[HTML]{EFEFEF}v1.4                  & \cellcolor[HTML]{EFEFEF}0.7789/ 0.7137 & \cellcolor[HTML]{EFEFEF}33.42/ 50.56 & \cellcolor[HTML]{EFEFEF}97.97                  \\
                                         & v1.5                                          & 0.5831/ 0.5648                         & 33.64/ 43.95                         & 92.80                                           & v1.5                                          & 0.7740/ 0.7143                         & 33.75/ 49.85                         & 98.10                                          \\
                                         & v2.1                                          & 0.5854/ 0.5655                         & 35.62/ 51.25                         & 99.29                                          & v2.1                                          & 0.7894/ 0.7010                         & 32.63/ 56.34                         & 99.80                                          \\
                                         \cmidrule{2-9}
                                         & v1.4                                          & 0.5828/ 0.5725                         & 35.62/ 43.92                         & 89.42                                          & v1.4                                          & 0.7789/ 0.7113                         & 33.42/ 53.47                         & 97.53                                          \\
                                         & \cellcolor[HTML]{EFEFEF}v1.5                  & \cellcolor[HTML]{EFEFEF}0.5831/ 0.5668 & \cellcolor[HTML]{EFEFEF}33.64/ 44.02 & \cellcolor[HTML]{EFEFEF}95.91                  & \cellcolor[HTML]{EFEFEF}v1.5                  & \cellcolor[HTML]{EFEFEF}0.7740/ 0.7092 & \cellcolor[HTML]{EFEFEF}33.75/ 53.44 & \cellcolor[HTML]{EFEFEF}98.83                  \\
                                         & v2.1                                          & 0.5854/ 0.5621                         & 35.62/ 50.90                         & 99.72                                          & v2.1                                          & 0.7894/ 0.7079                         & 32.63/ 56.66                         & 99.97                                          \\
                                         \cmidrule{2-9}
                                         & v1.4                                          & 0.5828/ 0.5761                         & 35.62/ 41.97                         & 85.72                                          & v1.4                                          & 0.7789/ 0.7149                         & 33.42/ 44.97                         & 96.63                                          \\
                                         & v1.5                                          & 0.5831/ 0.5678                         & 33.64/ 42.54                         & 93.23                                          & v1.5                                          & 0.7740/ 0.7198                         & 33.75/ 42.08                         & 96.43                                          \\
\multirow{-9}{*}{LoRA}                   & \cellcolor[HTML]{EFEFEF}v2.1                  & \cellcolor[HTML]{EFEFEF}0.5854/ 0.5680 & \cellcolor[HTML]{EFEFEF}35.62/ 45.84 & \cellcolor[HTML]{EFEFEF}98.95                  & \cellcolor[HTML]{EFEFEF}v2.1                  & \cellcolor[HTML]{EFEFEF}0.7894/ 0.7067 & \cellcolor[HTML]{EFEFEF}32.63/ 45.08 & \cellcolor[HTML]{EFEFEF}99.73                  \\
\midrule
                                         & \cellcolor[HTML]{EFEFEF}v1.4                  & \cellcolor[HTML]{EFEFEF}0.6946/ 0.6536 & \cellcolor[HTML]{EFEFEF}28.39/ 54.67 & \cellcolor[HTML]{EFEFEF}97.05                  & \cellcolor[HTML]{EFEFEF}v1.4                  & \cellcolor[HTML]{EFEFEF}0.7648/ 0.6985 & \cellcolor[HTML]{EFEFEF}27.86/ 62.58 & \cellcolor[HTML]{EFEFEF}99.67                  \\
                                         & v1.5                                          & 0.6940/ 0.6375                         & 30.91/ 47.67                         & 95.08                                          & v1.5                                          & 0.7638/ 0.7054                         & 27.72/ 60.86                         & 99.43                                          \\
                                         & v2.1                                          & 0.7407/ 0.6906                         & 29.66/ 72.92                         & 98.52                                          & v2.1                                          & 0.7417/ 0.6937                         & 29.42/ 52.87                         & 93.03                                          \\
                                         \cmidrule{2-9}
                                         & v1.4                                          & 0.6946/ 0.6597                         & 28.39/ 47.16                         & 93.45                                          & v1.4                                          & 0.7648/ 0.6997                         & 27.86/ 61.20                         & 98.93                                          \\
                                         & \cellcolor[HTML]{EFEFEF}v1.5                  & \cellcolor[HTML]{EFEFEF}0.6940/ 0.6414 & \cellcolor[HTML]{EFEFEF}30.91/ 47.58 & \cellcolor[HTML]{EFEFEF}96.74                  & \cellcolor[HTML]{EFEFEF}v1.5                  & \cellcolor[HTML]{EFEFEF}0.7638/ 0.6985 & \cellcolor[HTML]{EFEFEF}27.72/ 65.16 & \cellcolor[HTML]{EFEFEF}99.53                  \\
                                         & v2.1                                          & 0.7407/ 0.7008                         & 29.66/ 62.02                         & 98.06                                          & v2.1                                          & 0.7417/ 0.6878                         & 29.42/ 56.32                         & 93.83                                          \\
                                         \cmidrule{2-9}
                                         & v1.4                                          & 0.6946/ 0.6423                         & 28.39/ 41.45                         & 87.48                                          & v1.4                                          & 0.7648/ 0.6979                         & 27.86/ 50.68                         & 99.43                                          \\
                                         & v1.5                                          & 0.6940/ 0.6419                         & 30.91/ 44.94                         & 93.11                                          & v1.5                                          & 0.7638/ 0.6991                         & 27.72/ 53.27                         & 99.80                                          \\
\multirow{-9}{*}{DreamBooth}             & \cellcolor[HTML]{EFEFEF}v2.1                  & \cellcolor[HTML]{EFEFEF}0.7407/ 0.6836 & \cellcolor[HTML]{EFEFEF}29.66/ 71.98 & \cellcolor[HTML]{EFEFEF}99.63                  & \cellcolor[HTML]{EFEFEF}v2.1                  & \cellcolor[HTML]{EFEFEF}0.7417/ 0.6899 & \cellcolor[HTML]{EFEFEF}29.42/ 48.55 & \cellcolor[HTML]{EFEFEF}95.43                 \\
\bottomrule 
\end{tabular}
}
\vspace{-2mm}
\caption{Main Results of the Proposed DUAW. Here we provide the CLIP-Score, IL-NIQE, and protection success rate (SR) to evaluate the quality of generation by customized SD wo/w the watermarked training data. Rows with gray background represent white-box scenarios (\emph{i.e.}, DUAW trained on this model); others are black-box setting for measuring the transferability of DUAW.}
\vspace{-4mm}
\label{tab:main tab}
\end{table*}

\paragraph{Evaluation Settings}

We select SD v1.4, SD v1.5, and SD v2.1 (base version) to evaluate our DUAW and follow common SD fine-tuning settings (refer to Appendix B).
To evaluate the performance of the proposed watermark in subject-driven generation, we follow the evaluation protocol of the DreamBooth dataset, using 25 testing prompts per subject. For each subject, we generated 4 images per prompt, resulting in a total of 3000 test images. For style-driven generation, we employ ChatGPT to generate 25 testing prompts, encompassing diverse subjects such as landscapes, figures, pets, still life, and architecture. For each prompt, we employ the fine-tuned SD model to generate 10 images per artist, resulting in 3000 test images.\\

\paragraph{Evaluation Metric} Since DUAW is designed to introduce obvious corruptions into the images generated by the fine-tuned SD models, we utilize 1) CLIP score~\cite{hessel2021clipscore} to evaluate the similarity between the generated and original images and 2) IL-NIQE score~\cite{zhang2015feature}, a widely used no-reference image quality assessment (IQA), to quantify the perceptual quality of an image. Additionally, we introduce a classifier to identify the fine-tuned SD outputs images learned from watermarked data. We report the ratio of successfully identified images as 3) success rate (SR).

\subsection{Main Results}

\subsubsection{Image Similarity and Quality}

We report the quantitative results of the proposed DUAW in Tab.~\ref{tab:main tab}, which demonstrate that the proposed DUAW can effectively reduce the CLIP-Score, indicating an evident decrease in the similarity between the generated images and the original images. These results imply that DUAW successfully disrupts the customization process. Note that CLIP-Score measures the semantic similarity of the images rather than intricate details such as textures, so a slight decrease in CLIP-Score corresponds to obvious distortions (refer to Appendix C). Furthermore, the notable increase in IL-NIQE suggests a significant decline in the naturalness of generated images, aligning with the obvious distortion we can see. Hence, both CLIP-Score and IL-NIQE results demonstrate that the DUAW can disrupt the SD fine-tuning and distort the generated images of customized SD models, thus protecting the copyrighted images.

\subsubsection{Classification Success Rate}

DUAW can cause obvious distortion in output images of customized SD trained on watermarked images, so we can easily identify them via a simple classifier. The classifier we use is vision transformer (ViT)~\cite{dosovitskiy2021image} with a binary classification header; the dataset contains images generated by customized SD v1.4, v1.5, and v2.1 trained with/without watermarks. As shown in Tab.~\ref{tab:main tab}, our classifier can detect distortions in over 90\% of the images in most cases, with an exceptionally high average protection success rate in the Subject-driven task.
 We also report the performance of our classifier in Tab.~\ref{tab:classifier} to show the protection capability of DUAW comprehensively. The results reveal that a simple ViT classifier can achieve remarkable accuracy, recall, and precision, effectively distinguishing between watermarked and clean output images.

\begin{table}[!ht]
\centering
\scalebox{0.9}{
\begin{tabular}{ccc}
\toprule
Accuracy (\%) & Recall (\%) & Precision (\%) \\
\midrule
97.30         & 95.67       & 99.04         \\
\bottomrule
\end{tabular}
}
\vspace{-2mm}
\caption{Quantitative Results of the Naive ViT Classifier.
}
\vspace{-4mm}
\label{tab:classifier}
\end{table}

\subsubsection{Qualitative Results}
As shown in the left part of Fig.~\ref{fig:wikiart_db}, the proposed DUAW introduces distortions in the output of the fine-tuned SD model trained on watermarked data, significantly degrading the visual quality and compromising the aesthetic appeal. Furthermore, for different prompts and different customization methods, DUAW can effectively and stably protect copyrighted images. Meanwhile, it can be observed that the images generated by the DreamBooth exhibit more pronounced distortions since that DreamBooth tunes more weights of the SD model and can keep more features from reference images. We also report the transferability of DUAW in the right part of Fig.~\ref{fig:wikiart_db}, which shows that DUAW can induce obvious distortion in the outputs generated by customized SD models of different versions, indicating that our DUAW has the potential to protect images from future SD models.

\subsection{Ablation Study}

\paragraph{Transferability on Different VAE}
Given the proposed DUAW is crafted against VAE, we explore the protection capability of DUAW on different VAE variants. We use three commonly used VAEs~\footnote{https://huggingface.co/stabilityai/sd-vae-ft-ema}, including the ``base'' version used in SD v1.4 and v1.5, ``ft-ema'' fine-tuned on the base version with exponential moving average (EMA)~\cite{yaz2018unusual} weights, and ``ft-mse'' fine-tuned with emphasis on MSE loss. These VAE models are all plug-and-play and applicable to any version of SD. Therefore, we applied each of the three VAEs on SD v2.1 and reported the results. Remarkably, our DUAW demonstrates robustness against these VAE variants. As shown in Tab.~\ref{tab:vae}, the watermark trained on ft-mse shows high protection effectiveness on SD models using the other two VAEs. It is reasonable to assume that our DUAW is also capable of defending against future versions of SD models and their corresponding VAEs (in Appendix F, we also test SD XL 1.0).

\begin{figure}[!ht]
  \centering
  \includegraphics[width=7cm]{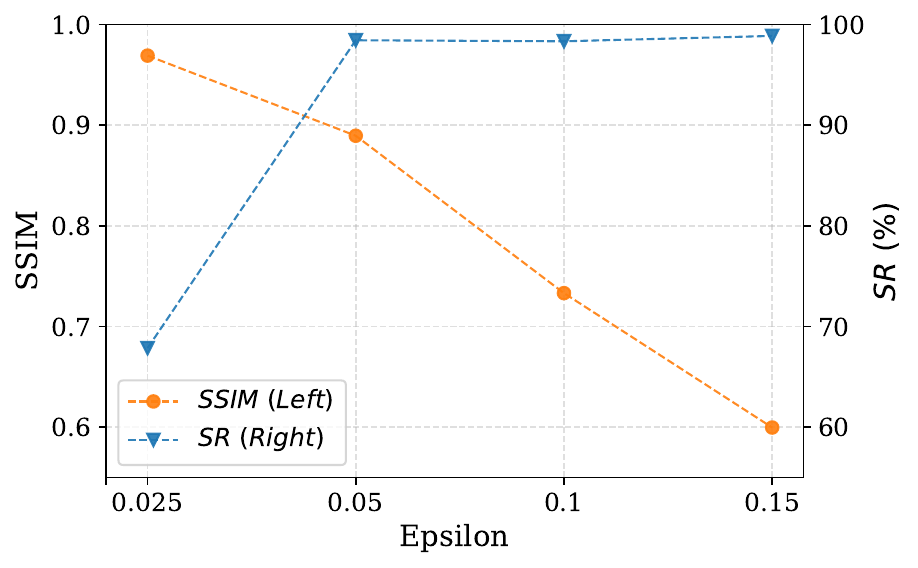}
  \vspace{-3mm}
  \caption{Experiment Results of Watermark Upper Bound. We provide the SSIM between the watermarked images and the original images, as well as the corresponding SR (\%) under different epsilon settings.}
  \vspace{-3mm}
  \label{fig:epsilon}
\end{figure}

\paragraph{Upper Bound of DUAW}
We investigate different upper bounds of the DUAW (\emph{i.e.}, epsilon) to compare the protective effectiveness of DUAW under different conditions. The visualizations of protected images using different epsilon values for DUAW are shown in Fig.~\ref{fig:epsilon}. Notably, at our default setting of 0.05, DUAW is scarcely discernible when added to the images, and its similarity to the original images remains relatively high. As epsilon increases, the distortions introduced by DUAW become progressively more pronounced, leading to better protection against SD models. However, this enhancement comes at the cost of compromising the watermark's invisibility. Trading off between SSIM and SR, we select epsilon=0.05 in our main experiments.

\begin{table}[!t]
\centering
\scalebox{0.8}{
\begin{tabular}{c|c|c|c}
\toprule
Method                       & VAE        & WikiArt & DreamBooth \\
\midrule
                             & base                           & 99.93                               & 98.70                                   \\
                             & ft-ema                         & 99.93                               & 99.33                                  \\
\multirow{-3}{*}{LoRA}       & \cellcolor[HTML]{EFEFEF}ft-mse & \cellcolor[HTML]{EFEFEF}99.05       & \cellcolor[HTML]{EFEFEF}99.90          \\
\midrule
                             & base                           & 99.77                               & 97.63                                  \\
                             & ft-ema                         & 99.47                               & 97.63                                  \\
\multirow{-3}{*}{DreamBooth} & \cellcolor[HTML]{EFEFEF}ft-mse & \cellcolor[HTML]{EFEFEF}99.85       & \cellcolor[HTML]{EFEFEF}97.90        \\
\bottomrule
\end{tabular}
}
\vspace{-2mm}
\caption{Transferability on Different Versions of VAE. We report the SR (\%) of DUAW here, which is trained on the ft-mse VAE and tested in different VAEs.}
\vspace{-4mm}
\label{tab:vae}
\end{table}

\begin{figure}[!ht]
  \centering
  \includegraphics[width=7cm]{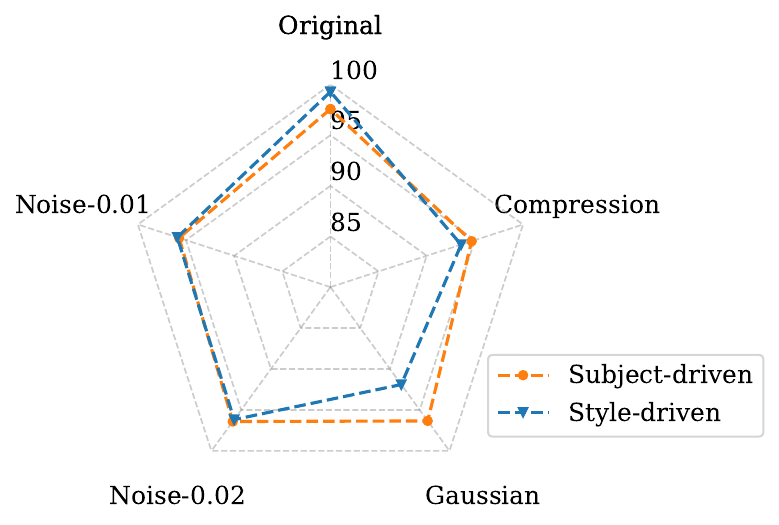}
  \vspace{-3mm}
  \caption{Robustness Analysis of DUAW. We use common image interference methods to perturb watermarked images and report SR (\%).}
  \vspace{-3mm}
  \label{fig:robust_vis}
\end{figure}

\paragraph{Robustness Analysis}
We explore the robustness of our watermark against common image interference methods, including random noise, Gaussian blur, and JPEG compression. As shown in Fig.~\ref{fig:robust_vis}, we observe that these interference methods marginally affect the protection effectiveness on SD v2.1, \emph{e.g.}, Gaussian blur and JPEG compression could slightly diminish distortions in the generated images. Nevertheless, the preservation success rate remains notably high, demonstrating the robustness of our DUAW against image interference.



\vspace{-3mm}

\section{Conclusion}
\label{sec:conclusion}

In this paper, we introduce DUAW, a data-free universal adversarial watermark designed to safeguard massive copyrighted images against various customization methods across different SD models. Based on our observation that VAE weights remain unchanged during SD fine-tuning and SD customization methods tend to preserve and amplify minor perturbations, we present a simple yet effective approach to create DUAW by perturbing the VAE component. To better protect the confidentiality of copyrighted images, we use LLM and pretrained SD models to generate a diverse training dataset. Our qualitative and quantitative experiments validate DUAW's efficacy in protecting copyrighted images and inducing pronounced distortion in the output images of customized SD models.


{\small
\bibliographystyle{ieee_fullname}
\bibliography{egbib}
}

\clearpage

\begin{appendices}

\section{Overview}
These appendices contain the following information:

\begin{itemize}

\item We give the detailed fine-tuning settings of DreamBooth and LoRA in Appendix B.

\item We give additional analysis regarding CLIP score as an evaluation metric in Appendix C.

\item We provide our surrogate training dataset along with the prompts used to generate said dataset in Appendix D.

\item We provide our evaluation dataset along with prompts used for style-driven generation in Appendix E.

\item We visualize the experiment results on SDXL 1.0 model in Appendix F.

\item We compare MS-SSIM and MSE as losses in training DUAW in Appendix G.


\end{itemize}

\section{Fine-tuning Settings}
\paragraph{Subject-driven generation}
For the DreamBooth method, we use a batch size of 1 and a learning rate of $5e^{-6}$ and fine-tuned the UNet of the SD model with 200 class images for 800 epochs. The instance prompt is set to "a photo of sks \textless class name\textgreater" (\emph{e.g.}, a photo of sks cat), and the class prompt is set to "a photo of \textless class name\textgreater" (\emph{e.g.}, a photo of cat).
As for the LoRA method, we employ the BLIP-2~\cite{li2023blip2} model to generate captions for all images and add the "sks" identifier to the captions (\emph{e.g.}, a photo of a sks dog in the grass). We fine-tune the SD model for 1500 epochs, with a batch size of 1 and a learning rate of $1e^{-4}$.\\
\paragraph{Style-driven generation}
For the DreamBooth method, we fine-tune the UNet and text-encoder of the SD model using 1000 class images for 800 epochs. The instance prompt is set to ``sks style'', and the class prompt is set to ``art style'';
For the LoRA method, we use the BLIP-2 model to generate captions for the images, adding the ``in sks style'' identifier to the captions. The fine-tuning of the SD model is performed for 1200 epochs.

\begin{figure}[t]
  \centering
  \includegraphics[width=8cm]{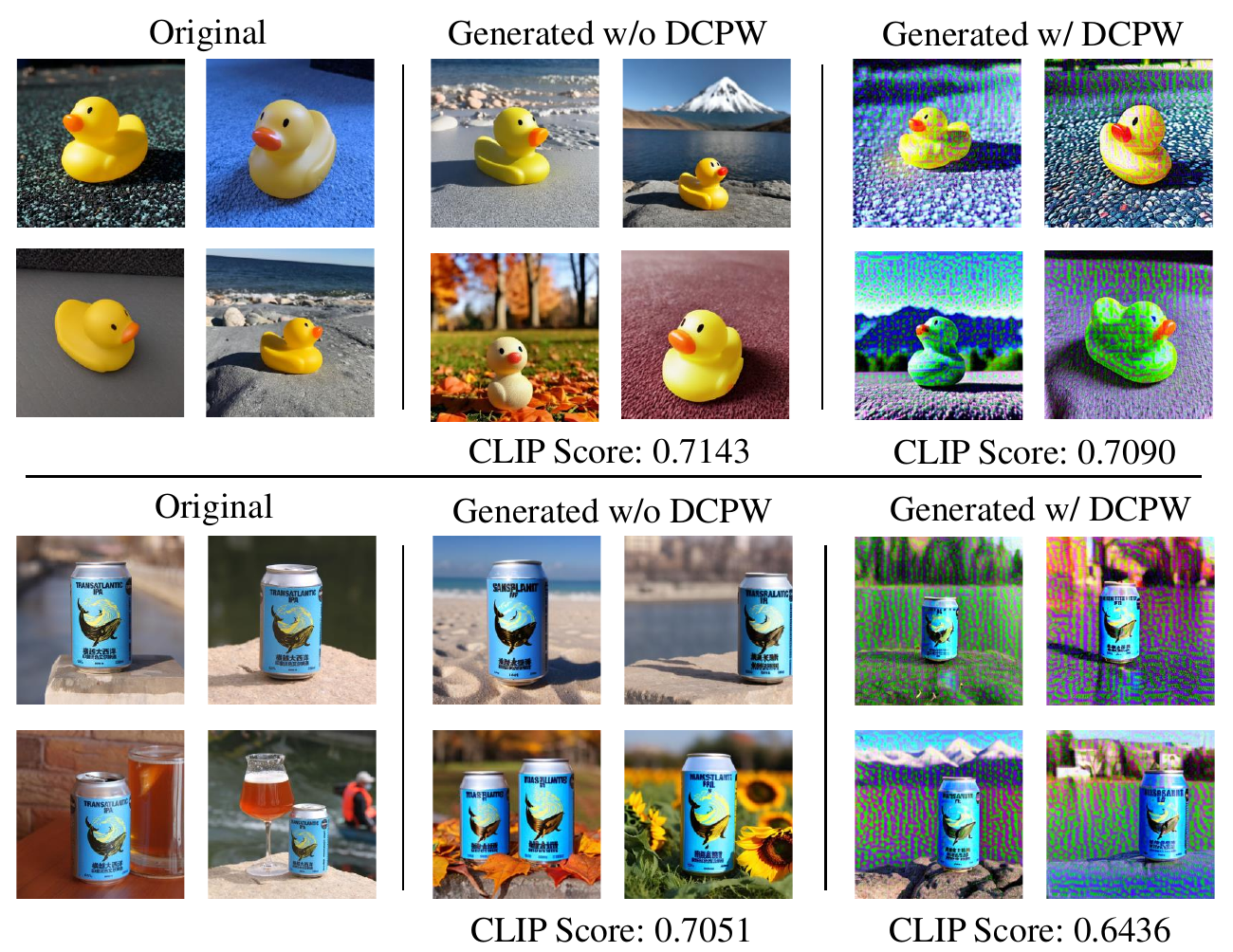}
  \vspace{-2mm}
  \caption{CLIP Score between Original and Generated Images with DUAW. It can be observed that even in the presence of highly noticeable distortions in the images, the CLIP score remains high.}
  \vspace{-2mm}
  \label{fig:clip_score}
\end{figure}

\section{CLIP score}
We use the CLIP score~\cite{hessel2021clipscore} metric as it is frequently used in evaluating subject and style-guided image generation. A decrease in the CLIP score indicates a drop in similarity between the generated and original images, thus indicating successful protection of copyrighted images. However, it is essential to note that the CLIP score is not a comprehensive measure of image quality, as it emphasizes semantic similarity rather than intricate details such as artifacts and distortions. Indeed, when strong distortions can be observed in the generated images, the CLIP score remains relatively high (Fig.~\ref{fig:clip_score}).

\begin{table*}[t]
\centering
\scalebox{0.8}{
\begin{tabular}{c|c}
\toprule
\textbf{Prompts} & \textbf{Styles} \\
\midrule
A quiet forest scene with a babbling brook and sunlight filtering through the trees.  &  Baroque  \\
\midrule
A lion standing on a sand dune.  &  Impressionism  \\
\midrule
An old man sleeping in a dimly lit room.  &  Surrealism  \\
\midrule
A deserted beach with crashing waves and a sunset in the distance. 
 &  Art Nouveau  \\
 \midrule
A pair of goldfish swimming in a spherical fish tank.  &  Post Impressionism  \\
\midrule
A surreal landscape with floating islands and a rainbow-colored sky. 
 &  Expressionism  \\
 \midrule
A quaint little cabin nestled in the woods, with a thatched roof and stone chimney.  &  Realism  \\
\midrule
A newlywed couple getting married in a beautiful church.  &  Renaissance  \\
\midrule
A craft room with a sewing machine, a work table, and a dog lying on a cushion.  &  Romanticism  \\
\midrule
A noblewoman sitting in a luxurious garden, surrounded by flowers and fountains.  &  Ukiyo-e  \\
\bottomrule
\end{tabular}
}
\vspace{-2mm}
\caption{Prompts and Styles for Surrogate Dataset Generation. We selected 10 prompts and 10 styles to generate the dataset, where each prompt can be combined with every style from the style zoo, resulting in 100 final prompts.}
\vspace{-2mm}
\label{tab:prompts and styles}
\end{table*}

\section{Surrogate Dataset}
\begin{figure}[t]
  \centering
  \includegraphics[width=6cm]{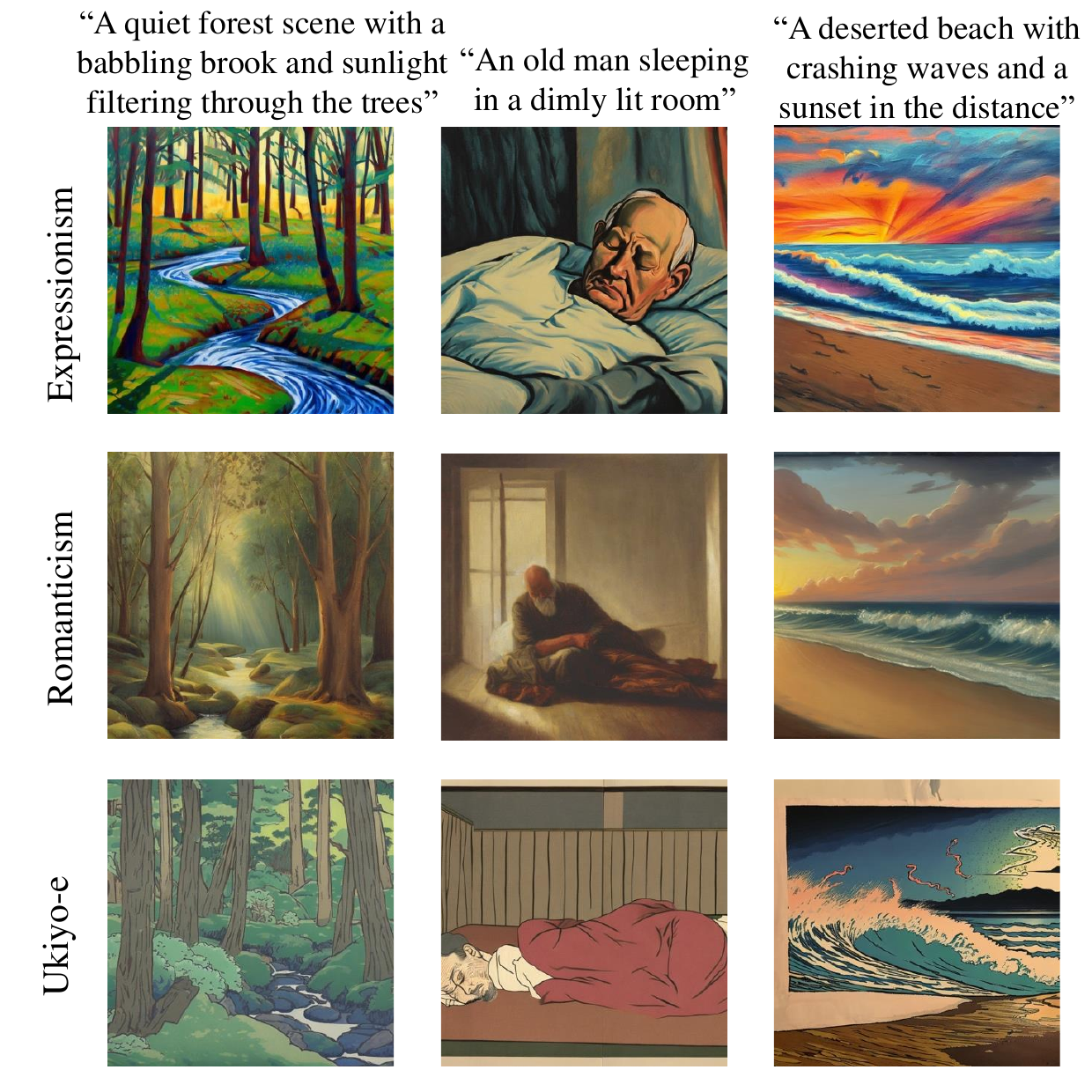}
  \vspace{-2mm}
  \caption{Images from the Surrogate Dataset. We generate diverse training images with prompts provided by ChatGPT and a style zoo.}
  \vspace{-2mm}
  \label{fig:training_data}
\end{figure}

To synthesize the surrogate dataset for the data-free scenario, we generate 10 prompts of painting contents using ChatGPT and combine each prompt with different artistic styles to generate images with high diversity. We use 10 distinct styles of ArtBench~\cite{liao2022artbench} to form the style zoo, which encompasses a wide range of artistic periods and genres. The prompts and styles are shown in Tab.~\ref{tab:prompts and styles}. Each prompt can combine with every available style to form a final prompt for SD generation. 
Utilizing SD v2.1, we generate a dataset containing 100 images. A subset of the dataset is showcased in Fig.~\ref{fig:training_data}.

\section{Evaluation Dataset and Prompts for Style-driven Generation}
\begin{figure}[t]
  \centering
  \includegraphics[width=8cm]{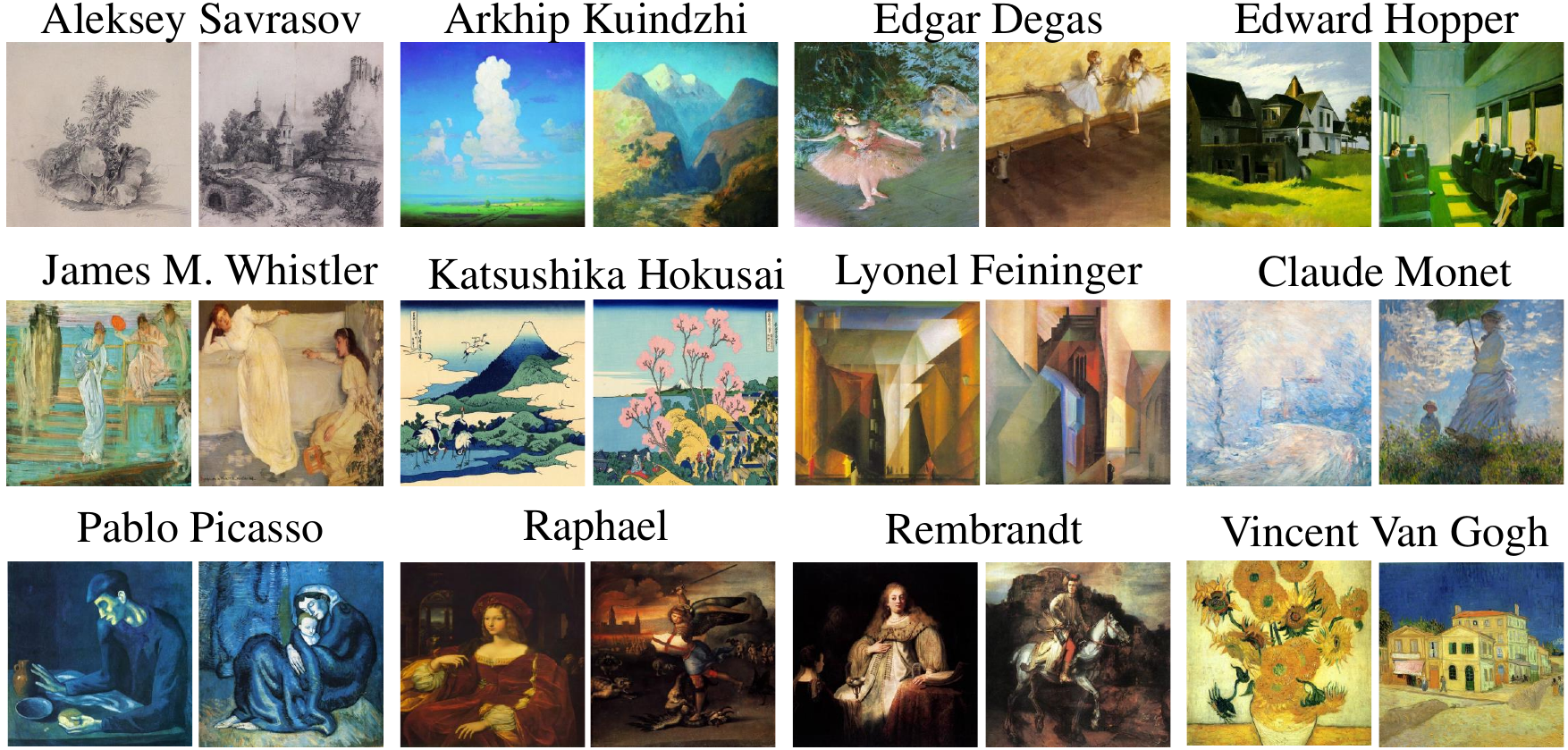}
  \vspace{-2mm}
  \caption{Selected Subset of WikiArt Dataset. We selected 12 artists, with each artist selecting 10 artworks, covering works from different periods and styles. These images are used as the evaluation dataset for style-driven generation.}
  \vspace{-2mm}
  \label{fig:wikiart_dataset}
\end{figure}

We choose a subset of the WikiArt dataset~\footnote{https://www.wikiart.org/}, a comprehensive collection of artworks from various artists, to evaluate DUAW on style-driven generation tasks. We selected 12 artists from the WikiArt dataset, each artist contributing 10 artworks with a resolution higher than $512\times512$. The selected artworks from the 12 artists are showcased in Fig.~\ref{fig:wikiart_dataset}. \\
To assess the effectiveness of our DUAW on this dataset, we conducted evaluations using 25 painting prompts generated by ChatGPT, which are listed in Tab.~\ref{tab:evaluation prompts}.

\begin{table}[t]
\centering
\scalebox{0.8}{
\begin{tabular}{c}
\toprule
\textbf{Evaluation prompt} \\
\midrule
A solitary tree standing against a sunset.\\
A serene beach with gentle waves and seashells.\\
A house in the snow.\\
A misty morning in a dense forest with towering trees.\\
A painting of some cloud floating in a clear blue sky.\\
A rustic wooden bridge over a calm river.\\
A playful kitten batting at a ball of yarn on a cozy blanket.\\
A full moon illuminating a peaceful nighttime cityscape.\\
A group of wildflowers blooming in a meadow.\\
A bear plushie sitting on a park bench.\\
A young girl with flowing curls and a lace-trimmed dress.\\
A charming cottage with a thatched roof and flower garden.\\
A city street with bustling cafes.\\
A field of sunflowers.\\
A night sky over a tranquil lake.\\
Snowy mountaintops.\\
A row of houses along a quaint cobblestone street.\\
A lady in a satin ballgown.\\
An antique pocket watch.\\
A peaceful garden with a trickling fountain and blooming roses.\\
A historic castle perched on a hill.\\
A serene lakeside cabin with a rowboat on the shore.\\
An adorable puppy playing with a chew toy in a sunlit room.\\
A delicate vase with a bouquet of colorful tulips.\\
An artist engrossed in creating a masterpiece.\\
\bottomrule
\end{tabular}
}
\vspace{-2mm}
\caption{Evaluation Prompts. We have selected 25 prompts to evaluate the attack effectiveness of DUAW, covering a variety of common scenarios.}
\vspace{-2mm}
\label{tab:evaluation prompts}
\end{table}

\section{Results on SDXL 1.0 Model}
\begin{figure}[t]
  \centering
  \includegraphics[width=8cm]{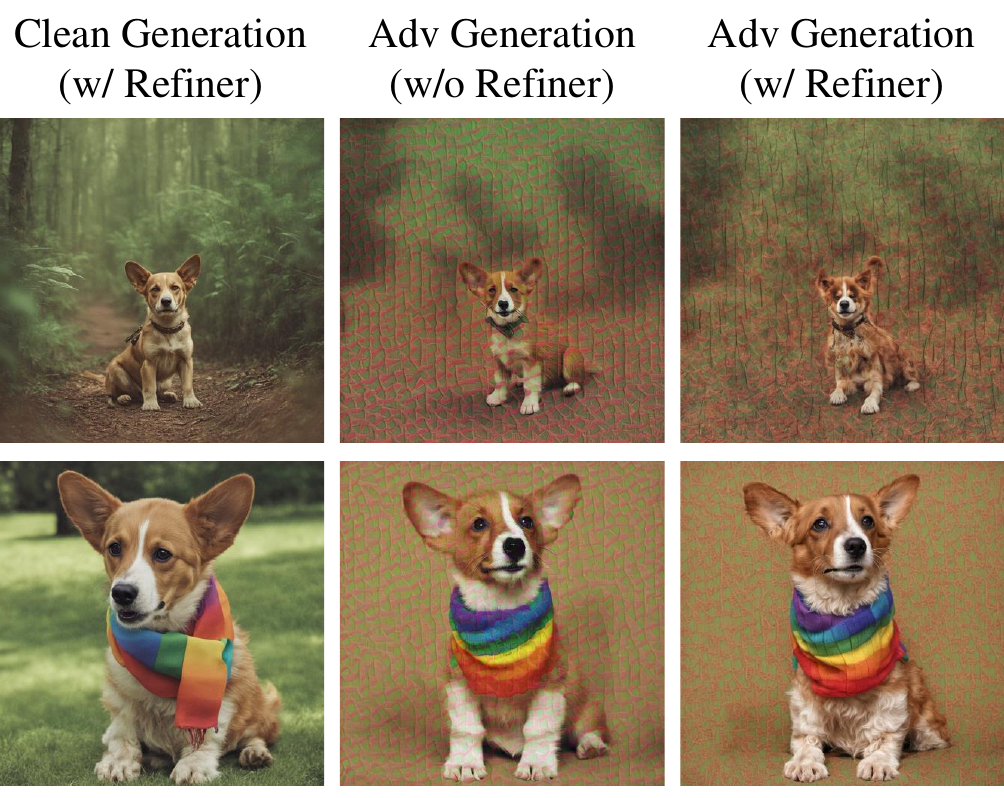}
  \vspace{-2mm}
  \caption{DUAW Results on SDXL 1.0 model. We present the clean generation w/o DUAW protection along with refined and unrefined generations with DUAW applied. Clearly, DUAW introduces distortions to the output of fine-tuned SDXL models, and subsequent refinement attempts fail to eliminate these distortions.}
  \vspace{-2mm}
  \label{fig:sdxl}
\end{figure}

Experiments show that our DUAW demonstrates strong transferability across various versions of SD models and their respective VAEs. We extend our experiments to include the SDXL 1.0 model~\cite{podell2023sdxl}, a recent variant specializing in producing high-resolution images with intricate details. SDXL employs a VAE trained from scratch and incorporates an additional image-to-image SD model, known as the Refiner, to enhance its outputs. By resizing the DUAW, initially trained on SD v2.1, to the size of $1024\times1024$, we use DreamBooth fine-tuning via LoRA~\footnote{huggingface.co/docs/diffusers/main/en/training/dreambooth} on the SDXL model for 500 epochs and visualize the results in Fig. ~\ref{fig:sdxl}. Notably, DUAW is capable of introducing noticeable distortions into SDXL's outputs, which remain unaffected by the Refiner model's attempts to remove them.

\begin{figure}[t]
  \centering
  \includegraphics[width=8cm]{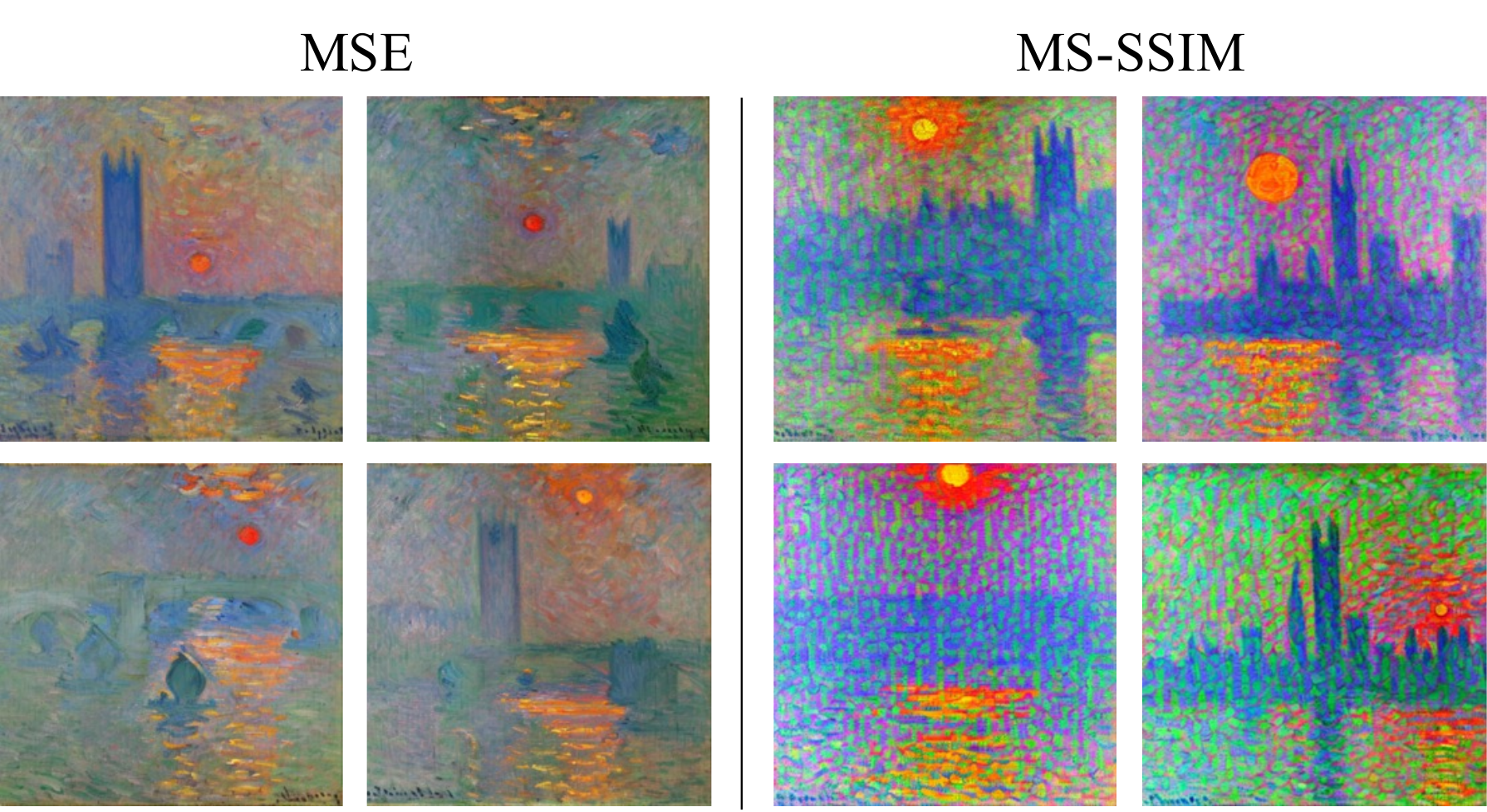}
  \vspace{-2mm}
  \caption{Comparison between MSE and MS-SSIM as Loss Functions. It is evident that when using MS-SSIM as loss, the outputs of VAE exhibit more noticeable distortions visible to the human eye.}
  \vspace{-2mm}
  \label{fig:mse_msssim}
\end{figure}

\section{MSE and MS-SSIM as Losses}
MSE and MS-SSIM are widely used metrics to assess image quality. MSE quantifies the average squared difference between the generated and reference images, while MS-SSIM accounts for structural similarity and perceptual quality, evaluating structural information at different scales and weighing their significance.

We train DUAW by employing MSE and MS-SSIM as loss function to measure the degree of distortions in VAE-decoded watermarked images and test the trained watermarks on SD v2.1 using DreamBooth. As shown in Fig.~\ref{fig:mse_msssim}, watermark trained with MS-SSIM yield superior protection performance compared to MSE, with the distortions in the images much more pronounced and the perceived image quality more degraded. This shows that MS-SSIM is an image similarity index closely aligned with human perception, and minimizing MS-SSIM can cause more noticeable distortions in the images than MSE.




\end{appendices}

\end{document}